\documentclass[letterpaper, 10 pt, conference]{ieeeconf}  

\IEEEoverridecommandlockouts                              

\usepackage{graphicx}
\usepackage{color, soul}
\usepackage{bbm}
\usepackage{amsmath,amssymb}
\usepackage{bm}

\DeclareMathOperator*{\argmax}{\arg\!\max}
\usepackage{hhline}
\usepackage{algorithm}
\usepackage{algpseudocode}
\usepackage{multirow}
\usepackage{array}
\usepackage{paralist}
\usepackage{verbatim}
\usepackage{url}
\usepackage{hyperref}
\usepackage{footnote}
\usepackage{hhline}

\def \hb {\bm{h}} 
\def \xb {\bm{x}} 
\def \db {\bm{d}} 
\def \fb {\bm{f}} 

\title{\LARGE \bf
 Learning Social Affordance Grammar from Videos:\\Transferring Human Interactions to Human-Robot Interactions
}

\algnewcommand\INPUT{\item[\algorithmicinput]}
\algnewcommand\algorithmicinput{\textbf{Input:}}

\author{Tianmin Shu$^{1}$, Xiaofeng Gao$^{2}$, Michael S. Ryoo$^{3}$ and Song-Chun Zhu$^{1}$
\thanks{$^{1}$Center for Vision, Cogntion, Learning, and Autonomy, University of California, Los Angeles, USA
        {\tt\small tianmin.shu@ucla.edu, sczhu@stat.ucla.edu}}%
\thanks{$^{2}$Department of Electronic Engineering, Fudan University, China
        {\tt\small xfgao13@fudan.edu.cn}}%
\thanks{$^{3}$School of Informatics and Computing, Indiana University, Bloomington, USA
        {\tt\small mryoo@indiana.edu}}
}

\begin{document}

\maketitle
\thispagestyle{empty}
\pagestyle{empty}

\begin{abstract}

In this paper, we present a general framework for learning social affordance grammar as a spatiotemporal AND-OR graph (ST-AOG) from RGB-D videos of human interactions, and transfer the grammar to humanoids to enable a real-time motion inference for human-robot interaction (HRI). Based on Gibbs sampling, our weakly supervised grammar learning can automatically construct a hierarchical representation of an interaction with long-term joint sub-tasks of both agents and short term atomic actions of individual agents. Based on a new RGB-D video dataset with rich instances of human interactions, our experiments of Baxter simulation, human evaluation, and real Baxter test demonstrate that the model learned from limited training data successfully generates human-like behaviors in unseen scenarios and outperforms both baselines.

\end{abstract}

\section{INTRODUCTION}


With the recent progress in robotics, robots now have been able to perform many complex tasks for humans. As a result, it is inevitable that the robots will interact with humans in various social situations, such as service robots taking care of elderly people, robot co-workers collaborating with humans in a workplace, or simply a robot navigating through human crowds. Similar to human social interactions, human-robot interactions (HRI) must also follow certain social etiquette or social norms, in order to make humans comfortable.

Conventional robot task planing only consider the effectiveness and efficiency of performing specific tasks, such as manufacturing, cleaning, and other activities that do not consider human values or preference. However, as J. J. Gibson pointed out, ``The richest and most elaborate affordances of the environment are provided by ... other people.'' \cite{Gibson}. A robot should reason the intention and feeling of humans who are near it and only perform socially appropriate actions while trying to achieve its own goal.

Therefore, in this paper, we focus on learning social affordances in human daily activities, namely action possibilities following basic social norms, from human interaction videos. More specifically, we are interested in the following three general types of human-robot interactions that we believe are most dominant interactions for robots: i) social etiquette, e.g., greeting, ii) collaboration, e.g., handing over objects, and iii) helping, e.g., pulling up a person who falls down. In addition, we also aim at developing a real-time motion inference to enable natural human-robot interactions by transferring the social affordance grammar. 

\begin{figure}[t!]
    \centering
    \includegraphics[trim={5 0 0 0},clip,width = 0.80\linewidth]{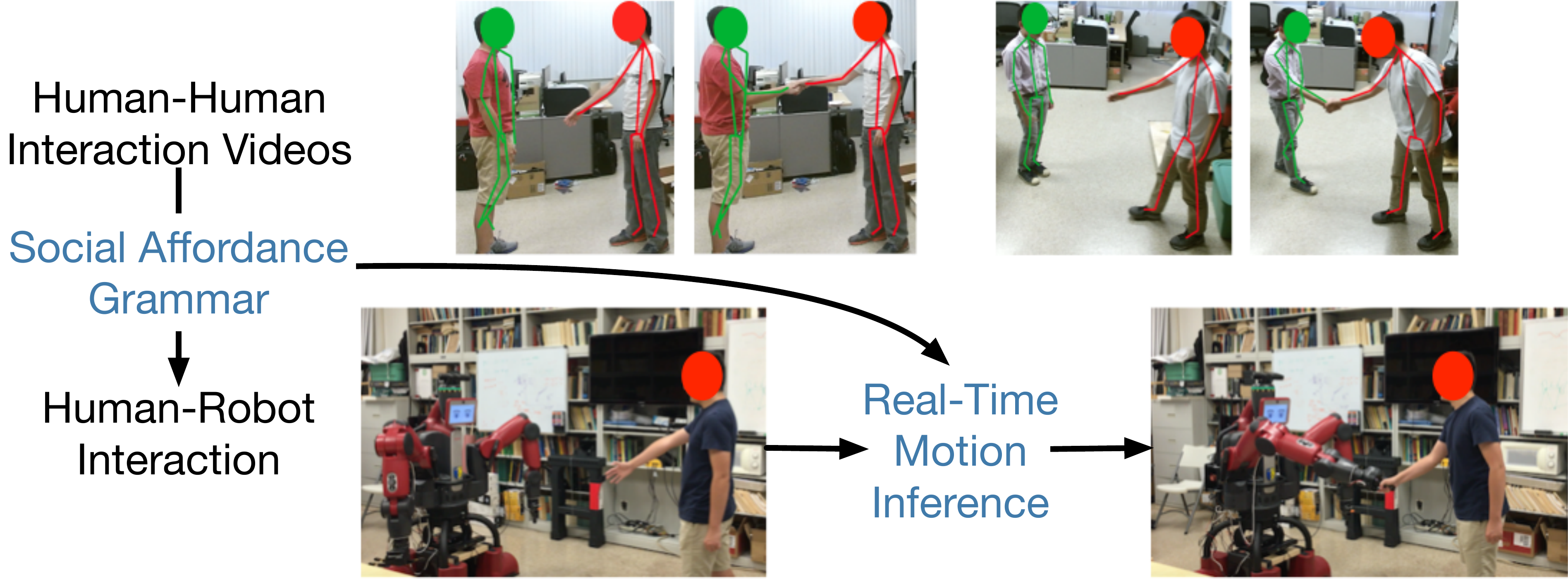}
    \vspace{-5pt}
    \caption{The framework of our approach.}
    \vspace{-10pt}
    \label{fig:intro}
\end{figure}

To this end, we propose a new representation for social affordances, i.e., social affordance grammar as a spatiotemporal AND-OR graph (ST-AOG), which encodes both important latent sub-goals for a complex interaction and the fine grained motion grounding such as human body gestures and facing directions. We learn the grammar from RGB-D videos of human interactions as Fig.~\ref{fig:intro} depicts. Our grammar model also enables short-term motion generation (e.g., raising an arm) for each agent independently while providing long-term spatiotemporal relations between two agents as sub-goals to achieve for both of them (e.g., holding the right hand of each other), which simultaneously maximizes the flexibly of our motion inference (single agent action) and grasps the most important aspects of the intended human-robot interactions (sub-goals in joint tasks).


\textbf{Contributions}:
\vspace{-3pt}
\begin{enumerate}
    \item A general framework for weakly supervised learning of social affordance grammar as a ST-AOG from videos;
    \item A real-time motion inference based on the ST-AOG for transferring human interactions to HRI.
\end{enumerate}


\section{RELATED WORK}


\textbf{Affordances}. In the existing affordance research, the domain is usually limited to object affordances \cite{Montesano2008,Kjellstrom2011,Bogdan2012,Zhu2014,Koppula2014,Pieropan2014,Sheng2015,Zhu2015}, e.g., possible manipulations of objects, and indoor scene affordances \cite{Gupta2011,Jiang2013}, e.g., walkable or standable surface, where social interactions are not considered. \cite{Shuijcai2016} is the first to propose a social affordance representation for HRI. However, it could only synthesize human skeletons rather than control a real robot, and did not have the ability to generalize the interactions to unseen scenarios. We are also interested in learning social affordance knowledge, but emphasize on transferring such knowledge to a humanoid in a more flexible setting.

\textbf{Structural representation of human activities}. In recent years, several structural representations of human activities for the recognition purposes have been proposed for human action recognition \cite{Gupta2009,Brendel2011,Pei2013,Lan2015} and for group activity recognition \cite{RyooIJCV2011,LanPAMI2012,Amer2012,Choi2014,Lan2014,Shu2015,Deng2016}. There also have been studies of robot learning of grammar models \cite{Lee2013, Yang2015, Xiong2016}, but they were not aimed for HRI.

\textbf{Social norms learning for robots}. Although there are previous works on learning social norms from human demonstrations aimed for robot planning, they mostly focused on relatively simple social scenarios, such as navigation \cite{Luber2012, okal2016}. On the contrary, we are learning social affordances as a type of social norm knowledge for much more complex interactions, which involve the whole body movements.



\section{FRAMEWORK OVERVIEW}

The framework of our approach illustrated in Fig.~\ref{fig:intro} can be outlined as follows:

\textbf{Human videos}. We collect RGB-D videos of human interactions, where human skeletons were extracted by Kinect. We use the noisy skeletons of these interactions as the input for the affordance learning.

\textbf{Social affordance grammar learning}. Based on the skeletons from human interaction videos, we design a Gibbs sampling based weakly supervised learning method to construct a ST-AOG grammar as the representation of social affordances for each interaction category.

\textbf{Real-Time motion inference}. For transferring human interactions to human-robot interactions, we propose a real-time motion inference algorithm by sampling parse graphs as hierarchical plans from the learned ST-AOG and generate human-like motion accordingly for a humanoid to interact with a human agent.


\section{REPRESENTATION}

   \begin{figure}[t!]
      \centering
      \includegraphics[trim={20 0 0 0}, clip, width = 0.85\linewidth]{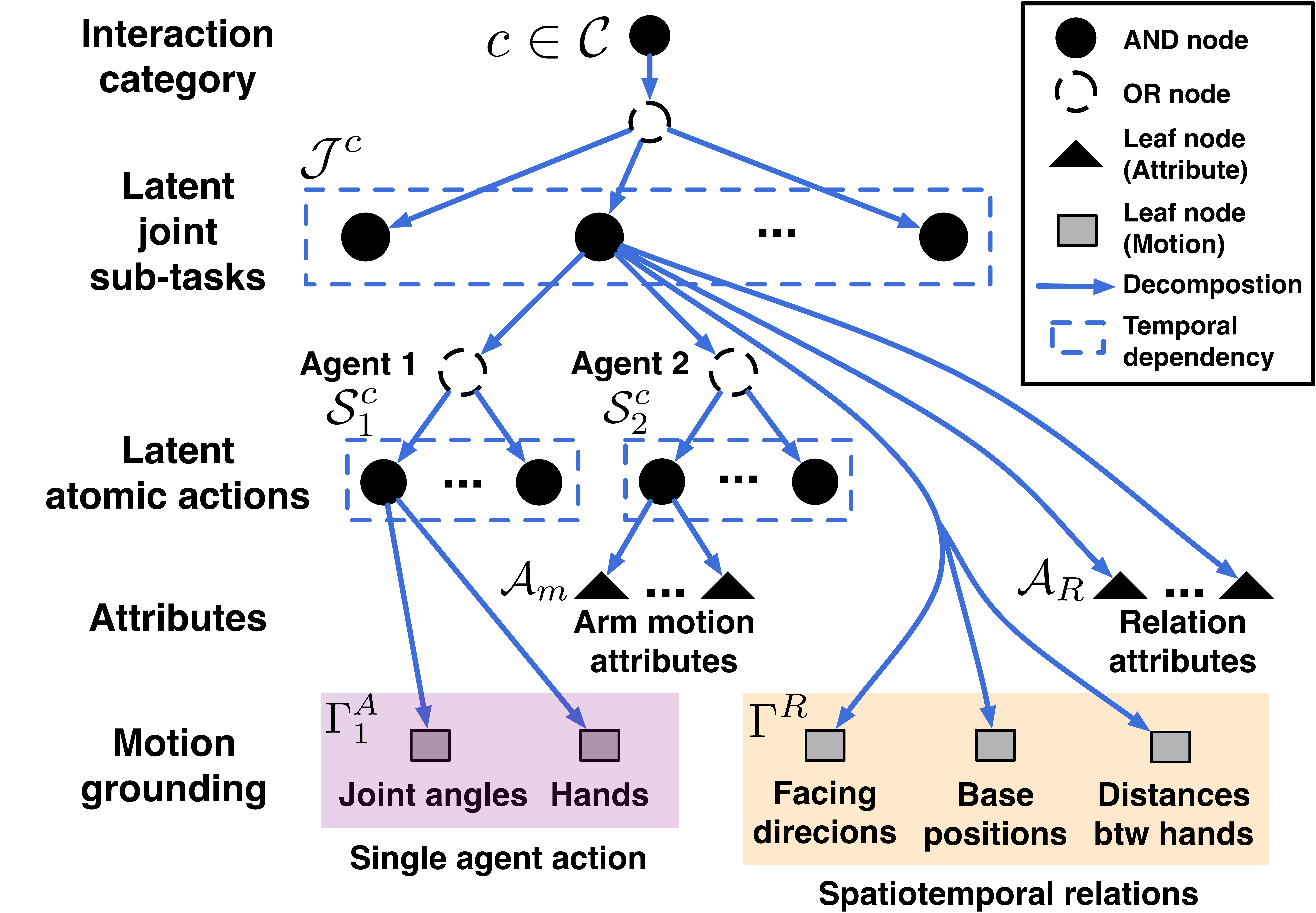}
      \vspace{-5pt}
      \caption{Social affordance grammar as a ST-AOG.}
      \vspace{-10pt}
      \label{fig:rep}
   \end{figure}
   
    \begin{figure}[t!]
      \centering
      \includegraphics[trim={10 5 5 5},clip,width = 0.45\linewidth]{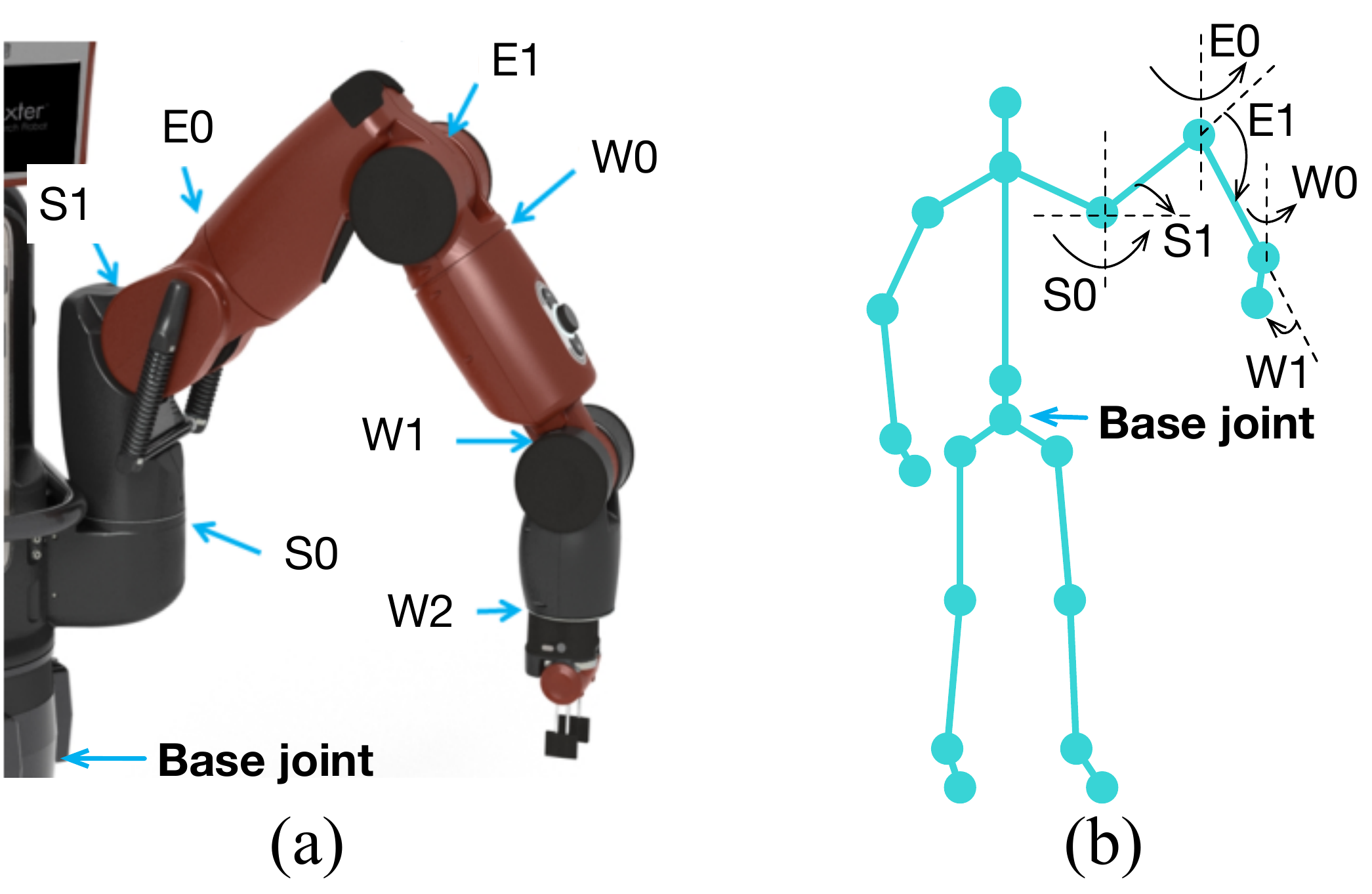}
      \vspace{-10pt}
      \caption[Caption for LOF]{(a) The joint angles of the arm of a Baxter robot (from \href{http://sdk.rethinkrobotics.com/wiki/Arms}{http://sdk.rethinkrobotics.com/wiki/Arms}), which are directly mapped to a human's arm (b). The additional angles (e.g., $w_2$) can be either computed by inverse kinematics or set to a constant value.}
      \vspace{-10pt}
      \label{fig:jointangles}
   \end{figure}
   
   \begin{figure*}[t!]
      \centering
      \includegraphics[width = 0.90\linewidth]{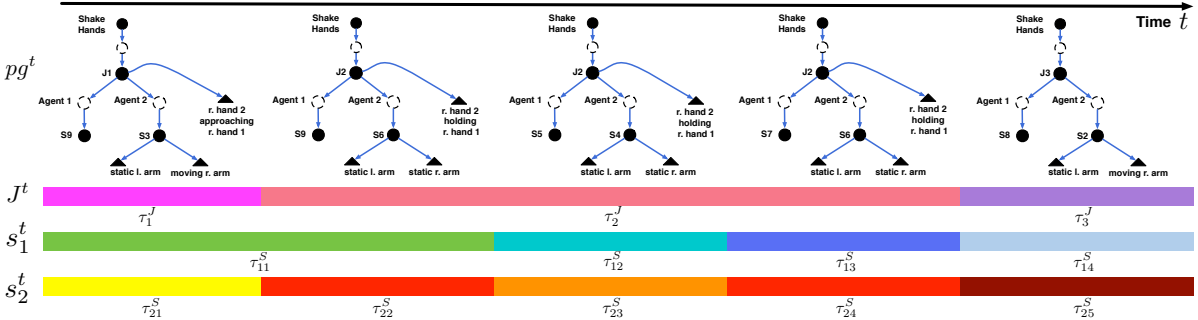}
      \vspace{-10pt}
      \caption{A sequence of parse graphs in a shaking hands interaction, which yields the temporal parsing of joint sub-tasks and atomic actions depicted by the colored bars (colors indicate the labels of joint sub-tasks or atomic actions).}
      \vspace{-10pt}
      \label{fig:pg}
   \end{figure*}

   We represent the social affordance knowledge as stochastic context sensitive grammar using a spatiotemporal AND-OR graph (ST-AOG), as shown in Fig.~\ref{fig:rep}. The key idea is to model the joint planning of two agents on top of independent action modeling of individual agents. Following the Theory of Mind (ToM) framework, a ST-AOG defines the grammar of possible robotic actions (agent 2) at a specific moment given the observation of agent 1's actions as the belief, the joint sub-tasks as sub-goals, and the interaction category as the overall goal.
   
   We first define a few dictionaries for the grammar model encoding the key elements in the social affordances. We constrain the human-robot interactions in a set of categories $\mathcal{C}$. Dictionaries of arm motion attributes $\mathcal{A}_M$ and relation attributes $\mathcal{A}_R$ are specified and shared across all types of interactions. Also, for each category $c$, there are dictionaries of latent joint sub-tasks $\mathcal{J}^c$, latent atomic actions of agent $i$, $\mathcal{S}_i^c$, where $\mathcal{S}_i^c$ are shared by different joint sub-tasks within $c$. Note that joint sub-tasks and atomic actions are not predefined labels but rather latent symbolic concepts mined from human activity videos, which boosts the flexibility of our model and requires much less human annotation efforts.
   
   There are several types of nodes in our ST-AOG: An AND node defines a production rule that forms a composition of nodes; an OR node indicates stochastic switching among lower-level nodes; the motion leaf nodes show the observation of agents' motion and their spatiotemporal relations; attribute leaf nodes provide semantics for the agent motion and spatiotemporal relations, which can greatly improve the robot's behavior. In our model, we consider four arm motion attributes, i.e., \textit{moving left/right arm}, \textit{static left/right arm}, and the relation attributes include \textit{approaching} and \textit{holding} between two agents' hands (possibly an object).
   
   The edges $\mathcal{E}$ in the graph represent decomposition relations between nodes. At the top level, a given interaction category leads to a selection of joint sub-tasks as the sub-goal to achieve for the given moment. A joint sub-task further leads to the atomic action selection of two agents and can also be bundled with relation attributes. An atomic action encodes a consistent arm motion pattern, which may imply some arm motion attributes of agent 2 for the purpose of motion inference. Some of the nodes in the dashed box are connected representing the ``followed by'' relations between joint sub-tasks or atomic actions with certain transition probabilities.
   
   The motion grounding is designed for motion transfer from a human to a humanoid, which entails social etiquette such as proper standing distances and body gestures. As shown in Fig.~\ref{fig:jointangles}, the pose of a human arm at time $t$ can be conveniently mapped to a robot arm by four degrees: $\theta^t = \langle s_0, s_1, e_0, e_1 \rangle$. The wrist angles are not considered due to the unreliable hand gesture estimation from Kinect. Thus, in an interaction whose length is $T$, there is a sequence of joint angles, i.e., $\Theta_{il} = \{\theta_{il}^t\}_{t = 1,\cdots,T}$ for agent $i$'s limb $l$, where $l = 1$ stands for left arm and $l = 2$ indicates right arm. Similarly the hand trajectories $H_{il} = \{\hb_{il}^t\}$ are also considered in order to have a precise control of the robot's hands. We model the spatiotemporal relations with agent 2's the relative facing directions, $O = \{o^t\}_{t = 1,\cdots,T}$, and relative base positions (in the top-down view), $X = \{\xb^t\}_{t = 1,\cdots,T}$, by setting the facing directions and base joint positions of agent 1 as references respectively. We also consider the distances between two agents' hands, $D_{ll^\prime} = \{\db_{ll^\prime}^t\}_{t = 1,\cdots,T}$ ($l$ is the limb of agent 1 and $l^\prime$ is the limb of agent 2) for the relations. The distances between agent 2's hands and an object can be included if an object is involved. For an interaction instance, we then define the action grounding of agent $i$ to be $\Gamma^A_i = \langle \Theta \rangle$, and the relation grounding of both agents to be $\Gamma^R = \langle O, X, D \rangle$, where $ \Theta = \{\Theta_{il}\}_{l=1,2}$, $H = \{H_{il}\}_{l=1,2}$, and $D = \{D_{ll^\prime}\}_{l,l^\prime \in \{1,2\}}$. Hence, the overall motion grounding is $\Gamma = \langle \{\Gamma^A_i\}_{i=1,2}, \Gamma^R \rangle$.
   
    Finally, the ST-AOG of interactions $\mathcal{C}$ is denoted by $\mathcal{G} = \langle \mathcal{C}, \{\mathcal{J}^c\}_{c \in \mathcal{C}}, \{\mathcal{S}_i^c\}_{c \in \mathcal{C}, i=1,2}, \mathcal{A}_M, \mathcal{A}_R, \Gamma,  \mathcal{E} \rangle$. At any time $t$, we use a sub-graph of the ST-AOG, i.e., a parse graph $pg^{t} = \langle c, j^t, s_1^t, s_2^t  \rangle$, to represent the actions of individual agents ($s_1^t$, $s_2^t$) as well as their joint sub-tasks ($j^t$) in an interaction $c$. Note that the attributes are implicitly included in the parse graphs since they are bundled with labels of $j^t$ and $s_2^t$. 
    
    For an interaction in $[1, T]$, we may construct a sequence of parse graphs $PG = \{pg^t\}_{t = 1,\cdots,T}$ to explain it, which gives us three label sequences: $J = \{j^t\}_{t = 1,\cdots,T}$, $S_1 = \{s_1^t\}$ and $S_2 = \{s_2^t\}$. By merging the consecutive moments with the same label of joint sub-tasks or atomic actions, we obtain three types of temporal parsing, i.e., $\mathcal{T}^J = \{\tau^J_k\}_{k = 1,\cdots,K^J}$, $\mathcal{T}_1^S = \{\tau^S_{1k}\}_{k = 1,\cdots,K_1^S}$, and $\mathcal{T}_2^S = \{\tau^S_{2k}\}_{k = 1,\cdots,K_2^S}$ for the joint sub-tasks and the atomic actions of two agents respectively, each of which specifies a series of consecutive time intervals where the joint sub-task or the atomic action remains the same in each interval. Hence, in $\tau_k^J = [t_k^1, t_k^2]$, $j^t = j(\tau_k^J)$, $\forall t \in \tau_k^J$, and for agent $i$, $s_i^t = s_i(\tau_{ik}^S)$, $\forall t \in \tau_{ik}^S$ in $\tau_{1k}^S = [t_{ik}^1, t_{ik}^2]$. Fig.~\ref{fig:pg} shows an example of the temporal parsing from the parse graph sequence. Note the numbers of time intervals of these three types of temporal parsing, i.e., $K^J$, $K_1^S$, and $K_2^S$, may be different. Such flexible temporal parsing allows us to model long-term temporal dependencies among atomic actions and joint sub-tasks. 

   
\section{PROBABILISTIC MODEL}

We propose a probabilistic model for our social affordance grammar model. 

Given the motion grounding, $\Gamma$, the posterior probability of a parse graph sequence $PG$ is defined as
\begin{equation}
 p(PG | \Gamma)\propto \underbrace{p(\{\Gamma_i^A\}_{i=1,2} | PG)}_{\text{arm motion likelihood}} \underbrace{p(\Gamma^R | PG)}_{\text{relation likelihood}} \underbrace{p(PG)}_{\text{parsing prior}}.
\label{eq:posterior}
\end{equation}

Conditioned on the temporal parsing of atomic actions and joint sub-tasks, the likelihood terms model the arm motion and the relations respectively, whereas the parsing prior models the temporal dependencies and the concurrency among joint sub-tasks and atomic actions. We introduce these three terms in the following subsections.

\subsection{Arm Motion Likelihood}

First, we define three types of basic potentials that are repeatedly used in the likelihood terms:

1) \textbf{Orientation potential} $\psi_o(\theta)$. This potential is a von Mises distribution of the orientation variable $\theta$. If $\theta$ has multiple angular variables, e.g., the four joint angles $\theta = \langle s_0, s_1, e_0, e_1\rangle$, then the potential is the product of the von Mises distributions of these individual angular variables.

2) \textbf{Three-dimensional motion potential} $\psi_{3v}(\xb)$. Assuming that spherical coordinate of $\xb$ is $(r,\theta,\phi)$, the potential is characterized by three distributions, i.e., $\psi_{3v}(\xb) = p(r)p(\theta)p(\phi)$, where the first one is a Weibull distribution and the remaining are von Mises distributions.

3) \textbf{Two-dimensional position potential} $\psi_{2v}(\xb)$. We fit a bivariate Gaussian distribution for $\xb$ in this potential. 

For joint angles and hand positions in an atomic action, we are interested in their final statuses and change during the atomic action. Thus, for the limb $l$ of agent $i$ in the interval $\tau^S_{ik}$ assigned with atomic action $s_i(\tau_{ik}^S) \in \mathcal{S^c}$ such that $s_i^t = s_i(\tau_{ik}^S)$, $\forall t \in \tau^S_{ik}$, the arm motion likelihood
\begin{equation}
\setlength{\arraycolsep}{0pt}
\begin{array}{lcl}
&&p(\Theta_{il}, H_{il} |  \tau^S_{ik}, s_i(\tau_{ik}^S))\\ &\propto&\underbrace{\psi_o(\theta_{il}^{t^\prime} - \theta_{il}^t)}_{\text{joint angles's change}} \underbrace{\psi_o( \theta_{il}^{t^\prime})}_{\text{final joint angles}} \underbrace{\psi_{3v}(\hb_{il}^{t^\prime} - \hb_{il}^{t})}_{\text{hand movement}} \underbrace{\psi_{3v}( \hb_{il}^{t^\prime})}_{\text{final hand position}},
\end{array}
\label{eq:motion_ll_limb}
\end{equation}
where $t = t_{ik}^1$ and $t^\prime = t_{ik}^2$ are the starting and ending moments of $\tau^S_{ik}$. Assuming independence between the arms, the arm motion likelihood for agent $i$ in $\tau_{ik}^S$ is
\begin{equation}
p(\Gamma_i^A| \tau^S_{ik}, s(\tau_{ik}^S) ) = \prod_l p(\Theta_{il}, H_{il} |  \tau^S_{ik}, s_i(\tau_{ik}^S)),
\label{eq:motion_ll_interval}
\end{equation}
and the arm motion likelihood for the entire interaction is
\begin{equation}
p(\Gamma_i^A| PG) = \prod_k p(\Gamma_i^A| \tau^S_{ik}, s(\tau_{ik}^S) ).
\label{eq:motion_ll_agent}
\end{equation}

Finally, the overall arm motion likelihood is the product of two agents' arm motion likelihood, i.e.,
\begin{equation}
p(\{\Gamma_i^A\}_{i=1,2} | PG) = \prod_i p(\Gamma_i^A | PG).
\end{equation}

\subsection{Relation Likelihood}

Relation likelihood models the spatiotemporal patterns hidden in facing directions $O$, base positions $X$, and the distances between two agents' hands during a joint sub-task. In a interval $\tau_k^J$ with the same joint sub-task label $j(\tau_k^J)$ such that $j^t = j(\tau_k^J)$, $\forall t \in \tau_k^J$, the relation likelihood is
\begin{equation}
\begin{array}{lcl}
\setlength{\arraycolsep}{0pt}
p(\Gamma^R | \tau_k^J, j(\tau_k^J)) &\propto& \displaystyle \underbrace{\psi_{o}(o^{t^\prime})}_{\text{facing direction}} \underbrace{\psi_{2v}(\xb^{t^\prime})}_{\text{base position}}\\
&& \displaystyle \cdot \prod_{l,l^\prime}\underbrace{\psi_{3v}(d_{ll^\prime}^{t^\prime})}_{\text{final hand distance}}\underbrace{\psi_{3v}(d_{ll^\prime}^{t^\prime} - d_{ll^\prime}^{t})}_{\text{distance change}},
\end{array}
\label{eq:relation_ll_interval}
\end{equation}
where $\tau^J_{k}$ starts at $t = t_k^1$ and ends at $t^\prime = t_k^2$.

Hence, the overall relation likelihood can be written as
\begin{equation}
p(\Gamma^R | PG) = \prod_k p(\Gamma^R | \tau_k^J, j(\tau_k^J)).
\end{equation}

\subsection{Parsing Prior}

The prior of a sequence of parse graphs is defined by the following terms:
\begin{equation}
\setlength{\arraycolsep}{0pt}
\begin{array}{lcl}
p(PG)&=& \underbrace{\prod_k p\left(|\tau_k^J| \mid j(\tau_k^J)\right)}_{\text{duration prior of joint sub-tasks}}\\
&& \cdot \underbrace{\prod_k p\left(|\tau_{1k}^S| \mid s_1(\tau_{1k}^S)\right) \prod_k p\left(|\tau_{2k}^S| \mid s_2(\tau_{2k}^S)\right)}_{\text{duration prior of atomic actions}}\\
&& \displaystyle \underbrace{\prod_{k > 1}p\left(s_1(\tau_{1k}^S) | s(\tau_{1k-1}^S)\right)}_{\text{action transition for agent 1}}\underbrace{\prod_{k > 1}p\left(s_2(\tau_{2k}^S) | s(\tau_{2k-1}^S)\right)}_{\text{action transition for agent 2}}\\
&& \cdot  \underbrace{\prod_{t}p(s_1^t | j^t)p(s_2^j | j^t)}_{\text{concurrency}} \underbrace{\prod_{k > 1}p\left(j(\tau_k^J) | j(\tau_{k-1}^J)\right)}_{\text{joint sub-task transition}},\\
\end{array}
\label{eq:prior}
\end{equation}
where the duration priors follow log-normal distributions and the remaining priors follow multinomial distributions.


\section{LEARNING}

The proposed ST-AOG can be learned in a weakly supervised manner, where we only specify the generic dictionaries of attributes and the sizes of the dictionaries of joint sub-tasks and atomic actions for each interaction. Given $N$ training instances, $\Gamma = \{\Gamma_n\}_{n = 1,\cdots,N}$, of an interaction category, where $\Gamma_n = \langle\{\Gamma^A_i\}_{i=1,2}, \Gamma^R_i \rangle$ is the motion grounding of instance $n$, the goal of learning is to find the optimal parsing graph sequence, $PG_i$, for each instance by maximizing the posterior probability defined in (\ref{eq:posterior}); then the ST-AOG is easily constructed based on the parse graphs.

It is intractable to search for the optimal parsing of atomic actions and joint sub-tasks simultaneously, which will take an exponential amount of time. Instead, we first 1) parse atomic actions for each agent independently and then 2) parse joint sub-tasks. Based on the likelihood distributions from the parsing results, we may 3) further obtain the implied attributes for each type of joint sub-tasks and atomic actions. We introduce the details in the rest of this section.

\begin{figure}[t!]
    \centering
    \includegraphics[trim={110 340 140 355},clip,width = 0.7\linewidth]{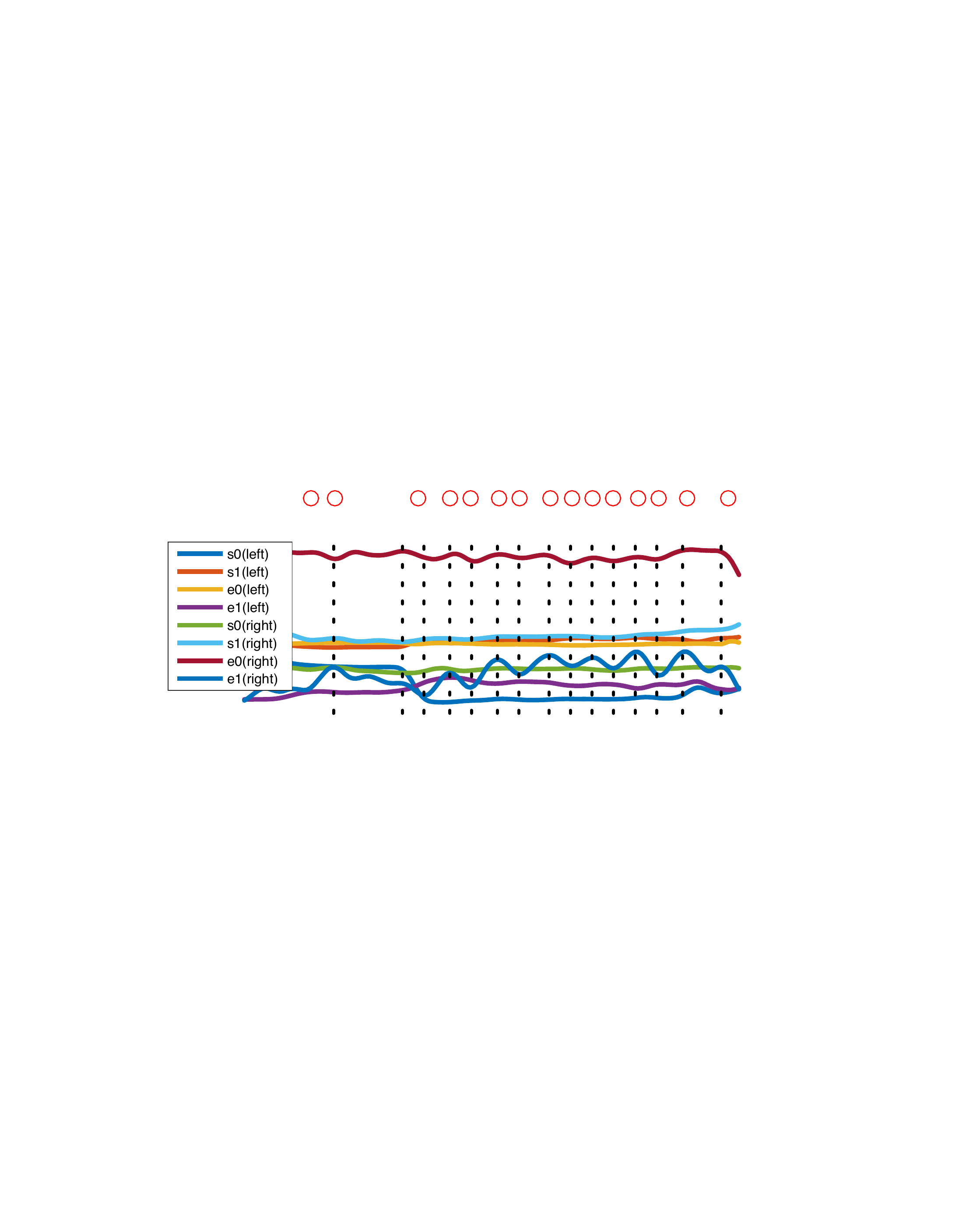}
    \vspace{-5pt}
    \caption{The curves show how the joint angles of agent 2's two arms change in an shaking hands interaction. The black dashed indicate the interval proposals from the detected turning points.}
    \vspace{-5pt}
    \label{fig:atomicparsing}
\end{figure}

\begin{figure*}[t!]
    \centering
    \includegraphics[width = 0.90\linewidth]{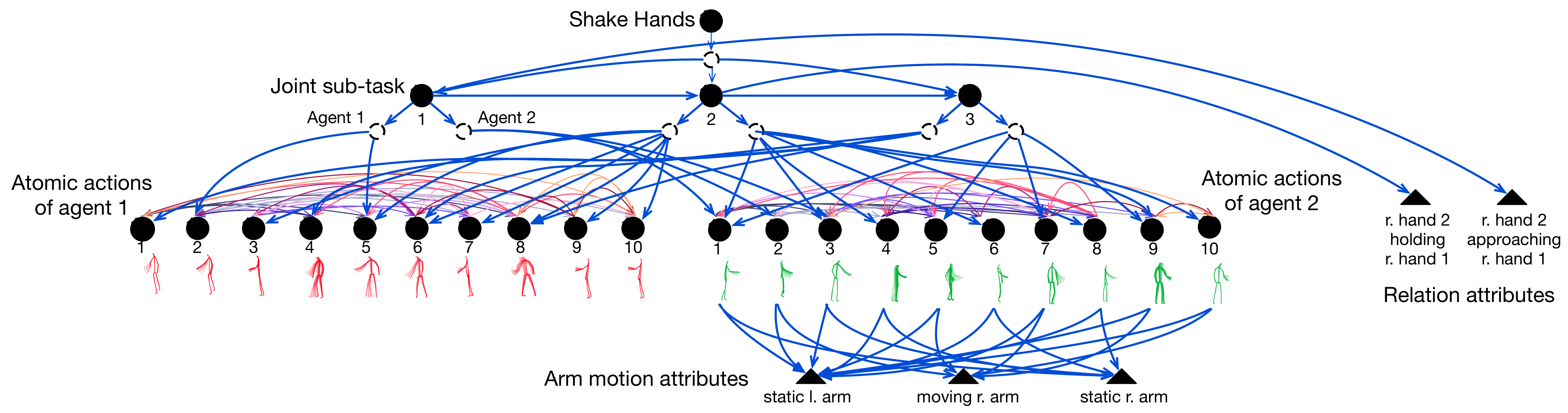}
    \vspace{-10pt}
    \caption{The learned ST-AOG for the \textit{Shake Hands} interaction (the motion grounding is not drawn in this figure due to the space limit). The numbers under AND nodes are the labels of joint sub-tasks or atomic actions. The edges between the atomic actions show the ``followed by'' temporal relations and their colors indicate which atomic actions are the edges' starting point. Similarly, the joint sub-tasks are also connected by edges representing the temporal dependencies between them. There is an example of each atomic actions from our training data, where the skeletons are overlaid with colors from light to dark to reflect the temporal order. The attributes that are not bundled to any atomic action or joint sub-task are not shown here.}
    \vspace{-5pt}
    \label{fig:AOG_Shake_Hands}
\end{figure*}

\subsection{Atomic Action Parsing}

We expect the motion in an atomic action to be consistent. Since the arm motion is characterized by joint angles and hand positions, the velocities of joints and hand movements should remain the same in an atomic action. Following this intuition, we propose the time intervals for the atomic actions of an agent by detecting the turning points of the sequences of joint angles (see Fig.~\ref{fig:atomicparsing}), which will naturally yields time intervals of atomic actions. To make the angles directly comparable, they are all normalized to the range of $[0, 1]$. 

To detect such turning points, we introduce a entropy function for a sequence $\{x^t\}$, i.e., $\mathcal{E}(t, w)$, where $t$ is the location of interest and $w$ is the window size. To compute $\mathcal{E}(t, w)$, we first count the histogram of the changes between consecutive elements, i.e., $x^t - x^{t-1}$ in the sub-sequence $\{x^{t^\prime}\}_{t^\prime=t-w,, t+w}$, and then $\mathcal{E}(t, w)$ is set to be the entropy of the histogram. By sliding windows with different sizes ($w = 2, 5, 10, 15$), we may detect multiple locations with entropy that is higher than a given threshold. By non-maximum suppression, the turning points are robustly detected.

After obtaining the time intervals, we assign optimal atomic action labels to each interval by Gibbs sampling. At each iteration, we choose an interval $\tau$ and sample a new label $s$ for it based on the following probability:
\begin{equation}
s \sim p(\Gamma_i^A \mid \tau, s)p(\tau, s \mid \mathcal{T}_i^S \backslash	 \tau, S_i \backslash \{s_i^t\}_{t \in \tau}).
\label{eq:action_sweep}
\end{equation}
Here, $p(\Gamma_i^A \mid \tau, s)$ is the likelihood in (\ref{eq:motion_ll_interval}), and based on the parsing prior in (\ref{eq:prior}), the labeling prior is computed as
\begin{equation}
p(\tau, s \mid \mathcal{T}_i^S \backslash \tau, S_i \backslash \{s_i^t\}_{t \in \tau}) = p(s \mid s^\prime) p(s^{\prime\prime} \mid s)p(|\tau| \mid s),
\end{equation}
where $s^\prime$ and $s^{\prime\prime}$ are the preceding and following atomic action labels in the adjacent intervals of $\tau$. If either of them is absent, the corresponding probability is then set to be 1. For each new label assignment, the parameters of the related likelihood and prior distributions should be re-estimated. To ensure the distinctness between adjacent intervals, $s$ can not be the same labels of the adjacent intervals. 

Therefore, after randomly assigning labels for the intervals as initialization, we conduct multiple sweeps, where in each sweep, we enumerate each interval and sample a new label for it based on (\ref{eq:action_sweep}). The sampling stops when the labeling does not change after the last sweep (convergence). In practice, the sampling can converge within $100$ sweeps coupled with a simulated annealing.

\subsection{Joint Sub-Task Parsing}

The joint sub-task parsing is achieved using a similar approach as atomic action parsing. We first propose the time intervals by detecting turning points based on the normalized sequences of $O$, $X$, and $D$. Then the labeling can also be optimized by a Gibbs sampling, where at each iteration, we sample a new joint sub-task label $j$ for an interval $\tau$ by
\begin{equation}
j \sim p(\Gamma^R \mid \tau, j)p(\tau, j \mid \mathcal{T}^J \backslash \tau, S^J \backslash \{j^t\}_{t \in \tau}),
\end{equation}
where $p(\Gamma^R \mid \tau, j)$ is defined in (\ref{eq:relation_ll_interval}) and the prior probability is derived from (\ref{eq:prior}) as
\begin{equation}
\setlength{\arraycolsep}{0pt}
\begin{array}{lcl}
&&p(\tau, j \mid \mathcal{T}^J \backslash \tau, S^J \backslash \{j^t\}_{t \in \tau})\\
&=&\displaystyle p(j \mid j^\prime)p(j^{\prime\prime} \mid j)p(|\tau| \mid j)\prod_{t\in \tau} p(s_1^t \mid j)p(s_2^t \mid j).
\end{array}
\label{eq:joint_sweep}
\end{equation}
Similar to (\ref{eq:action_sweep}), $j^\prime$ and $j^{\prime\prime}$ in the above prior probability are the preceding and following intervals' joint sub-task labels. The corresponding transition probability is assumed to be 1 if either of the adjacent interval does not exist. We also constrain $j$ to be different from the $j^\prime$ and $j^{\prime\prime}$ if they exist.

\subsection{Constructing ST-AOG}

After the previous two Gibbs sampling processes, the parameters of our probabilistic model are all estimated based on the parse graph sequences $\{PG_n\}_{n=1,\cdots,N}$. The ST-AOG of category $c$ is then constructed by the following three steps:
 
 \textbf{Initialization}. We start form a ``complete'' graph, where each non-leaf node is connected to all related lower level nodes (e.g., all joint sub-tasks, all atomic actions of the corresponding agent, etc.), except attribute leaf nodes. 
 
 \textbf{Edge removal}. Any edge between two joint sub-task nodes or two atomic action nodes is removed if it has a transition probability lower than a threshold (0.05). For each joint sub-task node, remove the edges connecting the OR node of agent $i$ to the atomic actions whose concurrency priors under the joint sub-task are lower than 0.1. Note that we use these thresholds for all interactions.
 
 \textbf{Attributes bundling}. Motion attributes: For each type of atomic action $s$ of agent $i$, a \textit{moving} attribute is bundled to a limb if the mean of the corresponding hand movement distribution specified in (\ref{eq:motion_ll_limb}) is lower than a threshold (we use 0.2 m in practice); otherwise, a \textit{static} attribute is bundled to the limb instead. Relation attributes: A type of joint sub-task will be associated with a \textit{holding} attribute between a pair of hands (or a hand and an object) if the mean final hand distance is lower than 0.15 m and the mean hand distance's change is lower than 0.1 m according to the corresponding distributions in (\ref{eq:relation_ll_interval}). If only the mean final hand distance meets the standard, an \textit{approaching} will be attached. For the case of multiple qualifying pairs for a hand, the one with the shortest mean distance is selected.

Fig.~\ref{fig:AOG_Shake_Hands} is a learned ST-AOG for \textit{Shake Hands} interactions. It can be seen that our learning algorithm indeed mines the critical elements of the interactions and clearly represents their relations through the structure of the ST-AOG.


\section{REAL-TIME MOTION INFERENCE}

If we replace agent 2 with a humanoid, we can therefore design a real-time motion inference enabling human-robot interaction based on the learned ST-AOG by sampling parse graphs and controlling the robot's motion accordingly.

For this, we propose two levels of inference procedures: 1) robot motion generation given the parse graphs, which is essentially transferring the socially appropriate motion from agent 2 in the grammar model to a humanoid; 2) parse graph sampling given the observation of the human agent's actions and the relation between the human agent and the robot according to the learned social affordance grammar.

\begin{algorithm}[t]\scriptsize
\caption{Parse Graph Sampling}
\label{alg:pgsampling}
\begin{algorithmic}[1]
\small

\INPUT The initial motion of two agents in $[1, T_0]$, i.e., $\Gamma(T_0)$

\State Infer $PG(T_0)$ by maximizing the posterior probability in (\ref{eq:posterior})

\State Let $t \leftarrow T_0 + 1$

\Repeat

    \State $\Gamma^\prime = \Gamma(t-1) \cup \{\theta_{1l}^t\}_{l=1,2} \cup \{h_{1l}^t\}_{l=1,2}$

    \State \parbox[t]{\dimexpr0.9\linewidth-\algorithmicindent}{Infer current atomic action of agent 1 by\newline $s_1^t = \argmax_{s} p(PG(t - 1) \cup \{s\} \mid \Gamma^\prime)$\strut}
    
    \ForAll{$j^t \in \mathcal{J}, s_2^t \in \mathcal{S}_2^c$ that are compatible with $s_1^t$}
        
        \State $pg^t \leftarrow \langle j^t, s_1^t, s_2^t \rangle$
        
        \State $PG(t) \leftarrow PG(t - 1) \cup \{pg^t\}$
        
        \State \parbox[t]{\dimexpr0.9\linewidth-\algorithmicindent}{Sample a new robot status at $t$, i.e., $\xb^t$, $o^t$, $\{\theta_{2l}^t\}$ and $\{\hb_{2l}^t\}$, as introduced in Sec.~\ref{sec:robotmotion} \strut}
        
        \State $\Gamma(t) \leftarrow \Gamma(t - 1) \cup \{\theta_{il}^t, \hb_{il}^t\}_{i,l=1,2} \cup \{\xb^t\} \cup \{o^t\}$
        
        \State Compute the posterior probability $p(PG(t) \mid \Gamma(t))$
        
    \EndFor
    
    \State \parbox[t]{\dimexpr0.95\linewidth-\algorithmicindent}{Choose the $pg^t$ and the corresponding new robot status that yield highest posterior probability to execute and update $PG(t)$ and $\Gamma(t)$ accordingly\strut}
    
    \State $t \leftarrow t + 1$
    
\Until{$t > T$}
\end{algorithmic}
\end{algorithm}

\subsection{Robot Motion Generation}
\label{sec:robotmotion}

As shown in Fig.~\ref{fig:jointangles}, we may use the motion grounding of agent 2 for the robot by joint mapping. The robot motion can be generated by sampling agent 2's base position $\xb^t$, facing direction (i.e., base orientation of the robot) $o^t$, joint angles $\{\theta_{2l}\}_{l=1,2}$, and hand positions (i.e., end effector positions) $\{\hb_{2l}^t\}_{l=1,2}$ at each time $t$ based on the motion history of agent 2, $\Gamma_2^A(t-1)$, and the spatiotemporal relations, $\Gamma^R(t-1)$, upon $t-1$ as well as the agent 1's motion, $\Gamma_1^A(t)$, and parse graphs, $PG(t) = \{pg^\tau\}_{\tau = 1,\cdots,t}$, upon $t$.

Since the arm motion is relative to the base position in our motion grounding, we first sample $\xb^t$ and $o^t$ w.r.t. the relative position and facing direction likelihood in (\ref{eq:relation_ll_interval}), the likelihood probabilities of which must be higher than a threshold ($0.05$ for $\xb^t$ and $0.3$ for $o^t$). To avoid jitter, we remain the previous base position and rotation if they still meet the criteria at $t$.

Then we update the joint angles for each robot arm. Without the loss of generality, let us consider a single arm $l\in\{1,2\}$. According to the atomic action $s_2^t$, we may sample desired joint angles $\hat{\theta}_{2l}^t$ and hand position $\hat{\hb}_{2l}^t$ w.r.t the corresponding likelihood terms in (\ref{eq:motion_ll_limb}). Since we do not model the wrist orientations, the desired $\hat{w}_0$, $\hat{w}_1$, $\hat{w}_2$ are always set to be 0 if the robot arm has these degrees of freedom (Fig.~\ref{fig:jointangles}a). If current joint sub-task entails an ``approaching'' or ``holding'' attribute for this limb, the desired hand position is set to the position of the target hand or object indicated by the attribute instead. To enforce the mechanical limits and collision avoidance, we minimize a loss function to compute the final joint angels $\theta_{il}^t$ for the robot arm:
\begin{equation}
\begin{array}{c}
\displaystyle \min_{\theta \in \Omega_{\theta}} \underbrace{\omega_h ||\fb_l(\theta)-\hat{\hb}_{2l}^t||_2^2}_{\text{hand position loss}} + \underbrace{\omega_a ||\theta - \hat{\theta}_{2l}^t||_2^2}_{\text{joint angle loss}} + \underbrace{\omega_s ||\theta - \theta_{il}^{t-1}||_2^2}_{\text{smoothness loss}},
\end{array}
\label{eq:loss}
\end{equation}
where $\fb_l(\theta)$ is the end effector position of $\theta$ based on the forward kinematics of the robot arm $l$; $\Omega_{\theta}$ is the joint angle space that follows the mechanical design (angle ranges and speed limits of arm joints) and the collision avoidance constraints, and $\omega_h$, $\omega_a$, $\omega_s$ are weights for the three types of loss respectively. By assigning different weights, we can design three control modes that are directly related to the attributes in ST-AOG:

1) \textbf{Hand moving mode}: if ``approaching'' or ``holding'' attributes are present in the current joint sub-task, we may use a larger $\omega_h$ to ensure an accurate hand position;

2) \textbf{Static mode}: if the first case does not hold and the atomic action has a ``static'' attribute for the limb, then $\omega_s$ should be much larger than $\omega_h$ and $\omega_a$; 

3) \textbf{Motion mimicking mode}: if none of the above two cases hold, we emphasize on joint angle loss (i.e., a large $\omega_a$) to mimic the human arm motion.


In practice, we set the large weight to be $1$ and the other two may range from $0$ to $0.1$.

\subsection{Parse Graph Sampling}

The Parse graph sampling algorithm is sketched in Alg.~{\ref{alg:pgsampling}}. The basic idea is to first recognize the action of agent 1. Then following the ST-AOG, we may enumerate all possible joint sub-tasks and atomic actions of agent 2 that are compatible with agent 1's atomic action, and sample a new robot status for each of them. Finally, we choose the one with the highest posterior probability to execute. Note that the facing direction of an agent is approximated by his or her moving direction (if not static) or the pointing direction of feet (if static).



\begin{table}[t!]\scriptsize
\caption{A summary of our new dataset (numbers of instances).}
\vspace{-10pt}
\label{table:dataset}
\begin{center}
\tabcolsep=0.08cm
\begin{tabular}{|c|c|c|c|c|c|c|}
\hline
Category & Scenario 1 & Scenario 2 & Scenario 3 & Scenario 4 & Total\\ \hline
Shake Hands & 19 & 10 & 0 & 0 & 29\\ \hline
High Five & 18 & 7 & 0 & 23 & 48\\ \hline
Pull Up & 21 & 16 &	9 & 0 & 46 \\ \hline
Wave Hands & 0 & 28 & 0 & 18 & 46\\ \hline
Hand Over & 34 & 6 & 8 & 7 & 55 \\
\hline
\end{tabular}
\end{center}
\vspace{-5pt}
\end{table}

\begin{figure*}[t!]
    \centering
    \includegraphics[trim={50 0 0 30},clip,width = 0.85\linewidth]{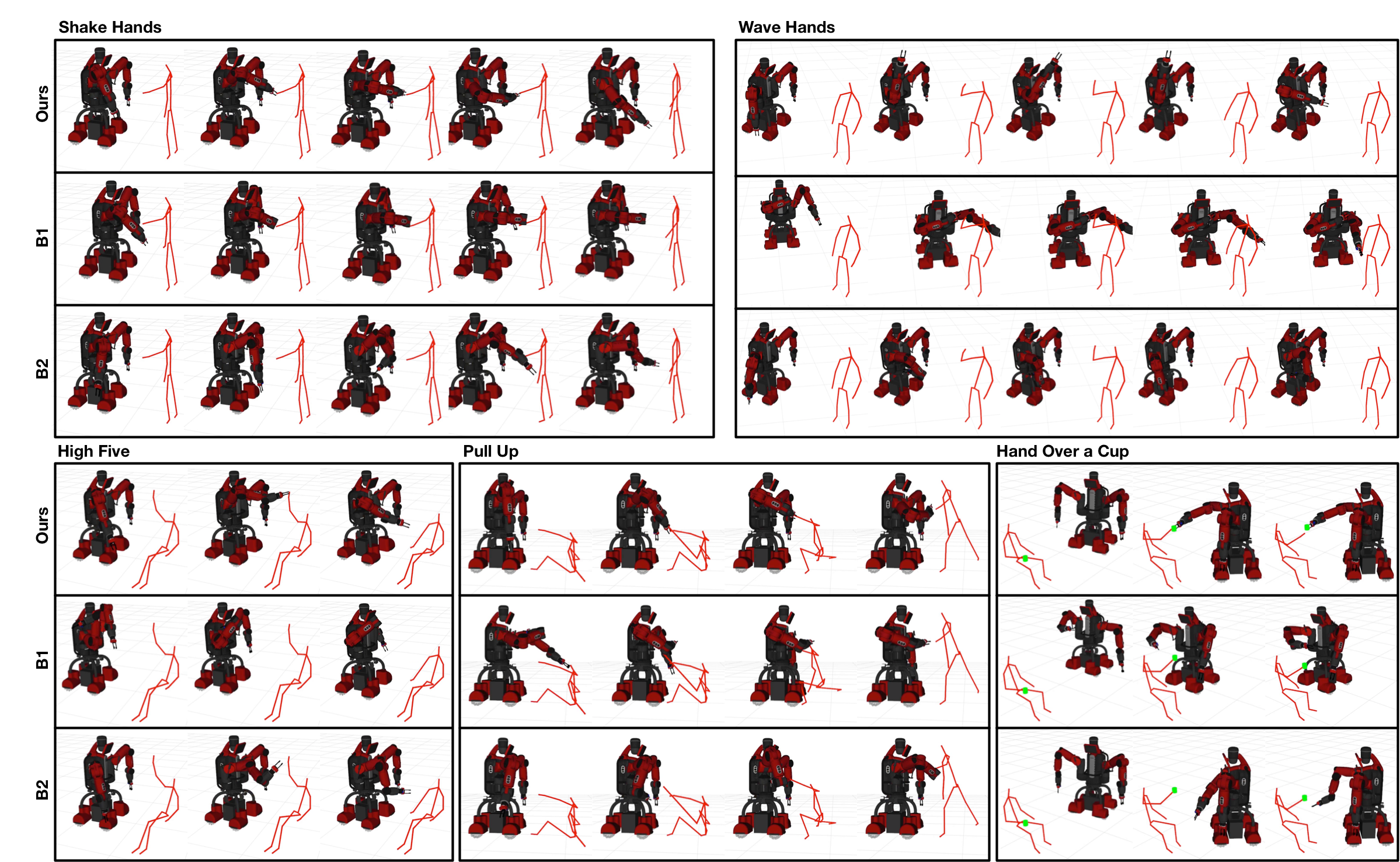}
    \vspace{-5pt}
    \caption{Qualitative results of our Baxter simulation.}
    \vspace{-10pt}
    \label{fig:simulation}
\end{figure*}

\begin{figure*}[t!]
    \centering
    \includegraphics[trim={0 0 0 0},clip,width = 1.0\linewidth]{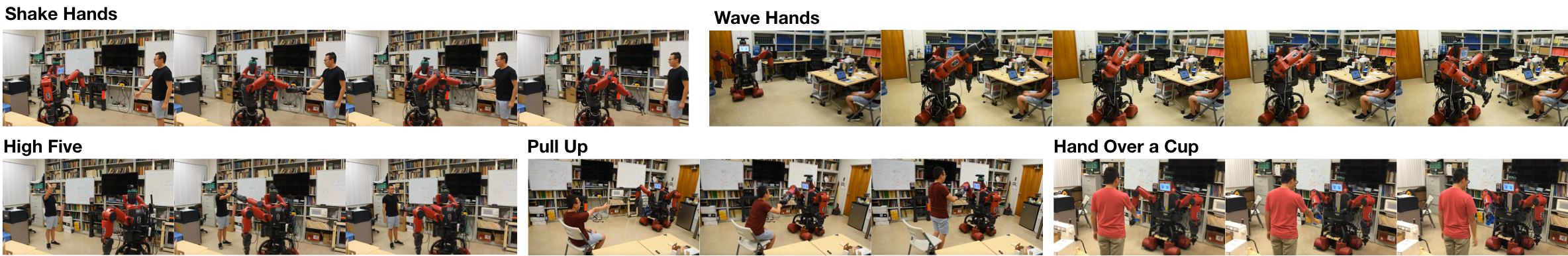}
    \vspace{-15pt}
    \caption{Qualitative results of the real Baxter test.}
    \vspace{-10pt}
    \label{fig:real}
\end{figure*}

\begin{table}[t!]\scriptsize
\caption{Mean joint angle difference (in radius degree) between the simulated Baxter and the ground truth skeletons.}
\label{table:quantitative}
\vspace{-5pt}
\centering
\tabcolsep=0.10cm
\begin{tabular}{c|c|c|c|c|c} \hline
Method   & Shake Hands & High Five & Pull Up & Wave Hands & Hand Over \\ \hline
B1 &   0.939 & 0.832 & 0.748 & 0.866 & 0.867    \\ \hline
B2  &   0.970 & 0.892 & 0.939 & 0.930 & 0.948   \\ \hline
Ours & \textbf{0.779} & \textbf{0.739} & \textbf{0.678} & \textbf{0.551} & \textbf{0.727} \\ \hline
\end{tabular}
\end{table}

\begin{table}[t!]\scriptsize
\caption{Human subjects' ratings of Baxter simulation generated by the three methods based on the two criteria.}
\label{table:ratings}
\vspace{-5pt}
\centering
\tabcolsep=0.08cm
\begin{tabular}{c|c|c|c|c|c|c} \hline
 & Source   & Shake Hands & High Five & Pull Up & Wave Hands & Hand Over \\ \hline
\multirow{3}{*}{Q1} & B1 & 3.22 $\pm$ 1.30 &  2.13 $\pm$ 1.09 & 2.75 $\pm$ 0.91 &  2.59 $\pm$ 1.20 & 2.19 $\pm$ 1.12    \\ \cline{2-7}
 & B2 & 2.14 $\pm$ 0.56 &  3.07 $\pm$ 1.22  & 2.11 $\pm$ 0.94 &  2.47 $\pm$ 0.69 & 1.48 $\pm$ 0.52   \\ \cline{2-7}
  & Ours & \textbf{4.45} $\pm$ 0.61 & \textbf{4.79} $\pm$ 0.41 & \textbf{4.53} $\pm$ 0.61 &  \textbf{4.82} $\pm$ 0.52 & \textbf{4.63} $\pm$ 0.53   \\ \hline
\multirow{3}{*}{Q2}  & B1 & 2.89 $\pm$ 0.99 &  2.38 $\pm$ 0.96  & 2.75 $\pm$ 0.55 &  2.00 $\pm$ 1.17 & 2.45 $\pm$ 0.71 \\ \cline{2-7}
 & B2 & 2.14 $\pm$ 0.83 & 2.93 $\pm$ 0.80 & 2.32 $\pm$ 1.00 & 1.60 $\pm$ 0.69 & 1.82 $\pm$ 0.63 \\ \cline{2-7}
  & Ours  & \textbf{4.20} $\pm$ 0.75 & \textbf{4.17} $\pm$ 0.62 & \textbf{4.25} $\pm$ 0.79 &  \textbf{4.65} $\pm$ 0.72 &   \textbf{3.97} $\pm$ 0.61   \\ \hline
\end{tabular}
\vspace{-5pt}
\end{table}

\section{EXPERIMENTS}

\textbf{Dataset}. There are two existing RGB-D video datasets for human-human interactions \cite{Yun2012, Shuijcai2016}, where the instances within the same category are very similar. To enrich the activities, we collected and compiled a new RGB-D video dataset on top of \cite{Shuijcai2016} using Kinect v2 as summarized in Table~\ref{table:dataset}, where \textit{Wave Hands} is a new category and the instances in scenario 1 of the other categories are from \cite{Shuijcai2016}. For \textit{Pull Up}, the first 3 scenarios are: A2 (agent 2) stands while A1 (agent 1) is sitting 1) on the floor or 2) in a chair; 3) A1 sits in a chair and A2 approaches. For the other categories, the four scenarios stand for: 1) both stand; 2) A1 stands and A2 approaches; 3) A1 sits and A2 stands nearby; 4) A1 sits and A2 approaches. In the experiments, we only use three fourths of the videos in scenario 1 (for \textit{Wave Hands}, it is scenario 2) as training data, and the remaining instances are used for testing. We plan to release the dataset.


\textbf{Baselines}. We compare our approach with two baselines adopted from related methods, extending these method further to handle our problem. The first one (B1) uses the method proposed in \cite{Shuijcai2016} to synthesize human skeletons to interact with the given human agent, from which we compute the desired base positions, joint angles and hand positions for the optimization method defined in (\ref{eq:loss}). Since \cite{Shuijcai2016} only models the end positions of the limbs explicitly and do not specify multiple modes as ours do, we use it with the weights of hand moving mode. The second baseline (B2) uses our base positions and orientations but solve the inverse kinematics for the two arms using an off-the-shelf planner, i.e., RRT-connect \cite{Kuffner2000} in MoveIt! based on the desired hand positions from our approach.

\subsection{Experiment 1: Baxter Simulation}
We first implement a Baxter simulation and compare the simulated robot behaviors generated from ours and the two baselines. For each testing instance, we give the first two frames of skeletons of two agents as the initialization; we then update the human skeleton and infer the new robot status accordingly at a rate of 5 fps in real-time. For \textit{Hand Over}, we assume that the cup will stay in the human agent's hand unless the robot hand is close to the center of the cup ($<$ 10 cm) for at least 0.4 s. Note that the planner in B2 is extremely slow (it may take more than 10 s to obtain a new plan), so we compute B2's simulations in an offline fashion and visualize them at 5 fps. Ours and B1 can be run in real-time.

Fig.~\ref{fig:simulation} shows a simulation example for each interaction. More results are included in the video attachment. From the simulation results, we can see that the robot behaviors (standing positions, facing directions and arm gestures) generated by ours are more realistic than the ones from baselines. Also, thanks to the learned social grammar, the robot can adapt itself to unseen situations. E.g., human agents are standing in the training data for ``High Five'', but the robot can still perform the interaction well when the human agent is sitting. 

We also compare the mean joint angle difference between the robot and the ground truth (GT) human skeletons (i.e., agent 2) captured from Kinect as reported in Table~{\ref{table:quantitative}}, which is one of the two common metrics of motion similarity \cite{Alibeigi2017} (the other one, i.e., comparing the end-effector positions, is not suitable in our case since humans and robots have different arm lengths). Although the robot has a different structure than humans', ours can still generate arm gestures that are significantly closer to the GT skeletons than the ones by baselines are. 

\subsection{Experiment 2: Human Evaluation}

To evaluate the quality of our human-robot interactions, we showed the simulation videos of three methods to 12 human subjects (UCLA students) who did not know that videos were from different methods. Subjects first watched two RGB videos of human interactions per category. Then for each testing instance, we randomly selected one method's simulation to a subject. The subjects only watched the assigned videos once and rated them based on two criteria: i) whether the purpose of the interaction is achieved (Q1), and ii) whether the robot's behavior looks natural (Q2). The ratings range from 1 (total failure/awkward) to 5 (successful/human-like).

The mean ratings and the standard deviations are summarized in Table~\ref{table:ratings}. Our approach outperforms the baselines for both criteria and has smaller standard deviations, which manifests its advantages on accurately achieving critical latent goals (e.g., holding hands) while keeping human-like motion. The rigid representation and failing to learn explicit hand relations affect B1's ability to adapt the robot to various scenarios. It also appears that only using a simple IK (B2) is probably insufficient: its optimization is only based on the current target position, which often generate a very long path and may lead to an awkward gesture. This makes the future target positions hard to reach as the target (e.g., a human hand) is constantly moving. 

\subsection{Experiment 3: Real Baxter Test}

We test our approach on a Baxter research robot with a mobility base. A Kinect sensor is mounted on the top of the Baxter's head to detect and track human skeletons. To compensate the noise from Kinect, we further take advantage of the pressure sensors on the ReFlex TakkTile Hand (our Baxter's right hand) to detect holding relations between the agents' hands. Although the arm movement is notably slower than the simulation due to the mechanical limits, the interactions are generally successful and reasonably natural. 

Since we only need joints on the upper body, the estimation of which is relatively reliable, the noisy Kinect skeletons usually do not greatly affect the control. In practice, temporal smoothing of the skeleton sequences is also helpful.


\section{CONCLUSIONS}
We propose a general framework of learning social affordance grammar as a ST-AOG from human interaction videos and transferring such knowledge to human-robot interactions in unseen scenarios by a real-time motion inference based on the learned grammar. The experimental results demonstrate the effectiveness of our approach and its advantages over baselines. In the future, it is possible to integrate a language model into the system to achieve verbal communications between robots and humans. In addition, human intention inference can also be added to the system.


\section*{ACKNOWLEDGMENT}
This work was funded by DARPA MSEE project FA 8650-11-1-7149 and ONR MURI project N00014-16-1-2007.


\bibliographystyle{IEEEtran}
\bibliography{IEEEabrv,icra17}

%
%

\end{document}